\pdfoutput=1
\documentclass[11pt,a4paper]{article}
\usepackage[hyperref]{naaclhlt2019}
\usepackage{times}
\usepackage{latexsym}

\usepackage{url}

\usepackage{graphicx}
\usepackage{todonotes}

\usepackage{algorithm}
\usepackage{varwidth}

\usepackage{tikz}
\usetikzlibrary{bayesnet}

\usepackage{algpseudocode}
\usepackage[fleqn]{amsmath}
\usepackage{amssymb}
\usepackage{mathtools}
\usepackage{multirow}
\usepackage{colortbl}
\usepackage[normalem]{ulem}
\usepackage[noabbrev,capitalise]{cleveref}
\algrenewcommand{\algorithmicrequire}{\textbf{Input:}}
\algrenewcommand{\algorithmicensure}{\textbf{Output:}}
\algnewcommand{\True}{\textbf{true}}
\algnewcommand{\False}{\textbf{false}}
\algnewcommand{\AAnd}{\textbf{ and }}
\algnewcommand{\Or}{\textbf{ or }}
\algnewcommand\algorithmicto{\textbf{to}}

\algdef{SE}{ForTo}{EndForTo}[3]%
    {\algorithmicfor\ $#1 = #2$ \algorithmicto\ $#3$}%
    {\algorithmicend\ \algorithmicfor}

\algdef{SE}{ForToProb}{EndForToProb}[3]%
    {\algorithmicfor\ $#1 = #2$ \algorithmicto\ $#3$}%
    {\algorithmicend\ \algorithmicfor}
\algtext*{EndForToProb}

\newcommand\E{\mathbb E}
\newcommand\mat[1]{\boldsymbol{\mathbf{#1}}}

\aclfinalcopy 

\title{Modelling Instance-Level Annotator Reliability for Natural Language Labelling Tasks}

\author{ Maolin Li\textsuperscript{\textnormal{1,2}}, 
         Arvid Fahlstr\"om Myrman\textsuperscript{\textnormal{2}},
         Tingting Mu\textsuperscript{\textnormal{2}}
         \textnormal{and} 
         Sophia Ananiadou\textsuperscript{\textnormal{1,2}}\\
  \textsuperscript{\textnormal{1}}National Centre for Text Mining, University of Manchester, United Kingdom \\
  \textsuperscript{\textnormal{2}}School of Computer Science, University of Manchester, United Kingdom \\
  {\tt \{maolin.li, arvidfm, tingting.mu, sophia.ananiadou\}@manchester.ac.uk}\\
 }
\date{}

\begin{document}
\maketitle
\begin{abstract}
When constructing models that learn from noisy labels produced by multiple annotators, it is important to accurately estimate the reliability of annotators. 
Annotators may provide labels of inconsistent quality due to their varying expertise and reliability in a domain. Previous studies have mostly focused on estimating each annotator's overall reliability on the entire annotation task.
However, in practice, the reliability of an annotator may depend on each specific instance. Only a limited number of studies have investigated modelling per-instance reliability and these only considered binary labels.
In this paper, we propose an unsupervised model which can handle both binary and multi-class labels. It can automatically estimate the per-instance reliability of each annotator and the correct label for each instance.
We specify our model as a probabilistic model which incorporates neural networks to model the dependency between latent variables and instances.
For evaluation, the proposed method is applied to both synthetic and real data, including two labelling tasks: text classification and textual entailment.
Experimental results demonstrate our novel method can not only accurately estimate the reliability of annotators across different instances, but also achieve superior performance in predicting the correct labels and detecting the least reliable annotators compared to state-of-the-art baselines.\footnote{Code is available at \url{https://github.com/createmomo/instance-level-reliability}}
\end{abstract}
\section{Introduction}
In many natural language processing (NLP) applications, the performance of supervised machine learning models depends on the quality of the corpus used to train the model.
Traditionally, labels are collected from multiple annotators/experts who are assumed to provide reliable labels.
However, in reality, these experts may have varying levels of expertise depending on the domains, and thus may disagree on labelling in certain cases \cite{aroyo2013measuring}.
A rapid and cost-effective alternative is to obtain labels through crowdsourcing \cite{snow2008cheap,poesio2013phrase,Poesio2017}.
In crowdsourcing, each instance is presented to multiple expert or non-expert annotators for labelling. However, labels collected in this manner could be noisy, since some annotators could produce a significant number of incorrect labels.
This may be due to differing levels of expertise, lack of financial incentive and interest \cite{Poesio2017}, as well as the tedious and repetitive nature of the annotation task \cite{raykar2010learning,bonald2017minimax}.

Thus, in order to ensure the accuracy of the labelling and the quality of the corpus, it is crucial to estimate the reliability of the annotators automatically without human intervention.

Previous studies have mostly focused on evaluating the annotators' overall reliability~\cite{Gurevych:2013:PWM:2484631,sheshadri2013square,Poesio2017}.
Measuring the reliability on a per-instance basis is however useful as we may expect certain annotators to have more expertise in one domain than another, and as a consequence certain annotation decisions will be more difficult than others. This resolves a potential issue of models that only assign an overall reliability to each annotator, where such a model would determine an annotator with expertise in a single domain to be unreliable for the model, even though the annotations are reliable within the annotator's domain of expertise. 

Estimating per-instance reliability is also helpful for unreliable annotator detection and task allocation in crowdsourcing, where the cost of labelling data is reduced using proactive learning strategies for pairing instances with the most cost-effective annotators~\cite{donmez2008proactive,li2017proactive}. Although reliability estimation has been studied for a long time, only a limited number of studies have examined how to model the reliability of each annotator on a per-instance basis. Additionally, these in turn have only considered binary labels ~\cite{yan2010modeling,yan2014learning,wang2017bi}, and cannot be extended to multi-class classification in a straightforward manner.

In order to handle both binary and multi-class labels, our approach extends one of the most popular probabilistic models for label aggregation, proposed by \citeauthor{hovy2013learning} \shortcite{hovy2013learning}. One challenge of extending the model is the definition of the label and reliability probability distributions on a per-instance basis. Our approach introduces a classifier which predicts the correct label of an instance, and a reliability estimator, providing the probability that an annotator will label a given instance correctly. The approach allows us to simultaneously estimate the per-instance reliability of the annotators and the correct labels, allowing the two processes to inform each other. 
Another challenge is to select appropriate training methods to learn a model with high and stable performance. We investigate training our model using the EM algorithm and cross entropy. For evaluation, we apply our method to six datasets including both synthetic and real-world datasets (see Section \ref{subsec:data}). In addition, we also investigate the effect on the performance when using different text representation methods and text classification models (see Section \ref{subsec:settings}).

Our contributions are as follows:
firstly, we propose a novel probabilistic model for the simultaneous estimation of per-instance annotator reliability and the correct labels for natural language labelling tasks.
Secondly, our work is the first to propose a model for modelling per-instance reliability for both binary and multi-class classification tasks.
Thirdly, we show experimentally how our method can be applied to different domains and tasks by evaluating it on both synthetic and real-world datasets.
We demonstrate that our method is able to capture the reliability of each annotator on a per-instance basis, and that this in turn helps improve the performance when predicting the underlying label for each instance and detecting the least reliable annotators.
\section{Related Work}
\subsection{Modelling Annotator Reliability}
\label{subsec:related_work_reliability}
Probabilistic graphical models have been widely used for inferring the overall reliability of annotators in the absence of ground truth labels.
Approaches include modelling a single overall reliability score for each annotator ~\cite{whitehill2009whose,welinder2010multidimensional,NIPS2011_4396,NIPS2012_4627,demartini2012zencrowd,hovy2013learning,rodrigues2014sequence,Li:2014:CAT:2735496.2735505,Li:2014:RCH:2588555.2610509}, estimating the reliability of each annotator on a per-category basis ~\cite{dawid1979maximum,NIPS2012_4490,pmlr-v22-kim12,zhang2014spectral}, and estimating the sensitivity and specificity for each annotator in binary classification tasks ~\cite{raykar2010learning}.

Fewer attempts have been made to model the per-instance reliability of annotators, focusing mainly on medical image classification. One approach is that by~\citeauthor{yan2010modeling}~\shortcite{yan2010modeling,yan2014learning} who use logistic regression to predict the per-instance reliability of annotators.
~\citeauthor{wang2017bi}~\shortcite{wang2017bi} used a modified support vector machine (SVM;~\citeauthor{cortes1995support}~\citeyear{cortes1995support}) loss, modelling the per-instance reliability as the distance from the given instance to a separation boundary.

\subsection{True Label Prediction in Crowdsourcing}
True label prediction in crowdsourcing is the aggregation of labels produced by different annotators to infer the correct label of each instance. Majority voting assigns to each instance the most commonly occurring label among the annotators, which 
can result in a high agreement between the predicted label and the ground truth for some NLP tasks~\cite{snow2008cheap}.
~\citeauthor{dawid1979maximum}~\shortcite{dawid1979maximum},
~\citeauthor{whitehill2009whose}~\shortcite{whitehill2009whose},
~\citeauthor{raykar2010learning}~\shortcite{raykar2010learning},
~\citeauthor{welinder2010multidimensional}~\shortcite{welinder2010multidimensional},
~\citeauthor{NIPS2012_4627}~\shortcite{NIPS2012_4627},
~\citeauthor{NIPS2012_4490}~\shortcite{NIPS2012_4490},
~\citeauthor{pmlr-v22-kim12}~\shortcite{pmlr-v22-kim12},
~\citeauthor{hovy2013learning}~\shortcite{hovy2013learning},
~\citeauthor{yan2010modeling} (\citeyear{yan2010modeling};~\citeyear{yan2014learning}),
~\citeauthor{Li:2014:RCH:2588555.2610509}~\shortcite{Li:2014:RCH:2588555.2610509} and
~\citeauthor{zhang2014spectral}~\shortcite{zhang2014spectral}
investigated binary or multi-class label prediction using probabilistic graphical models.~\citeauthor{NIPS2011_4396}~\shortcite{NIPS2011_4396}, ~\citeauthor{wang2017bi}~\shortcite{wang2017bi}, and  ~\citeauthor{bonald2017minimax}~\shortcite{bonald2017minimax} formalised the label prediction as an optimisation problem.
~\citeauthor{rodrigues2014sequence}~\shortcite{rodrigues2014sequence} and~\citeauthor{nguyen2017aggregating}~\shortcite{nguyen2017aggregating} investigated how to aggregate sequence labels using probabilistic graphical models.
\section{Methodology}
\subsection{Model}
\begin{figure}
\centering
\begin{tikzpicture}[scale=0.3]

  \node[obs]                               (a) {$a_{ij}$};
  \node[latent, above=of a, xshift=-1.2cm] (t) {$t_i$};
  \node[latent, above=of a, xshift=1.2cm]  (r) {$r_{ij}$};

  \edge {r,t} {a} ; %

  \plate {ra} {(r)(a)} {$M$} ;
  \plate {} {(r)(t)(a)(ra.north west)(ra.south west)(ra.east)} {$N$} ; 

\end{tikzpicture}
\caption{Graphical model.}
\label{fig:graphical_model}
\medskip
\centering
\small
\begin{varwidth}{\linewidth}
\begin{algorithmic}
  \ForToProb{i}{1}{N}
    \State $t_i \sim \mathrm{Categorical}(f_t(\mathbf{x}_i))$
    \ForToProb{j}{1}{M}
      \State $r_{ij} \sim \mathrm{Bernoulli}(f_r(\mathbf{x}_i, j))$
      \State $a_{ij} \sim \begin{cases}\mathrm{Uniform}(T) & r_{ij} = 0 \\ \delta_{t_i} & r_{ij} = 1\end{cases}$
    \EndForToProb
 \EndForToProb
\end{algorithmic}
\end{varwidth}
\caption{Generative process for our method.
$f_t$ is the classifier, returning a probability distribution over predicted labels, and $f_r$ is the reliability estimator, returning the probability that the annotator is accurate for the instance.
$\mathrm{Uniform}(T)$ is a uniform distribution over the categories in $T$.
$\delta_{t_i}$ is the deterministic distribution that only takes on the value $t_i$.}
\label{fig:generative_process}
\end{figure}

In the description of our model we let $N$ be the number of training instances, $M$ the number of annotators, $\mathbf x_i$ the $i$th training instance, $t_i$ its true underlying label, $T$ the set of values $t_i$ can take on, $r_{ij}$ whether annotator $j$ is reliable for the $i$th instance, and $a_{ij}$ the label that annotator $j$ gave the $i$th instance.
Below we describe the components of the model in more detail.

\paragraph{Probabilistic Model:} Our model is inspired by the method proposed by~\citeauthor{hovy2013learning}~\shortcite{hovy2013learning}, and it shares the same graphical representation (see Figure \ref{fig:graphical_model}).
The distributions of the model, however, are defined differently, as can be seen in Figure \ref{fig:generative_process}, due to the inclusion of a classifier and a reliability estimator.

We assume that the underlying label $t_i$ depends only on the corresponding instance, while the reliability $r_{ij}$ depends on the instance and the identity of the annotator.
If $r_{ij} = 0$, then the annotator $j$ is unreliable for instance $\mathbf x_i$, and a label is chosen randomly from among the available categories.
Otherwise, the annotation $a_{ij}$ is set to be the correct label.

\paragraph{Classifier:} The classifier $f_t(\mathbf x_{i})$ provides the predicted probabilities of an instance belonging to each category, $p(t_i \mid \mathbf{x}_i)$.
$t_i$ is the underlying label for instance $\mathbf x_i$, the $i$th instance, and takes a value in the set of categories $T$.
Note that there is no restriction on what classifier is used, other than that it can be trained using expectation maximisation.
The inclusion of a classifier directly in the model means that it can be trained while taking into account the uncertainty of the data and predictions, as opposed to first making a hard assignment of a label for each instance and training the classifier post-hoc.

\paragraph{Reliability Estimator:} The reliability estimator $f_r(\mathbf x_i, j)$ predicts the probability of annotator $j$ producing the correct label for instance $\mathbf x_i$, $p(r_{ij} \mid \mathbf{x}_i)$.
$r_{ij}$ is a binary variable, with $1$ and $0$ representing annotator $j$ being reliable and unreliable for instance $\mathbf x_i$, respectively.
The reliability estimator is modelled as a feed-forward neural network, where $j$ is encoded as a one-hot vector. The exact representation of $\mathbf{x}_i$ depends on the model used for the classifier. If the classifier is a neural network, the output of the last hidden layer is used; otherwise, the original feature vector is used. 
\subsection{Learning}
\label{subsec:learning}
\subsubsection*{Pre-training}
\label{subsubsec:initialisation}
As the number of parameters in our model is much larger than that of previous studies~\cite{yan2010modeling,yan2014learning,wang2017bi} due to the introduction of both a classifier and a reliability estimator, the model is much harder to train from scratch.
Therefore, before we start training the model, we first pre-train the classifier using labels predicted by a simpler method as targets, using e.g.\ majority voting or the method proposed by~\citeauthor{dawid1979maximum}~\shortcite{dawid1979maximum}. Although these labels may be noisy, we have observed empirically that a better initialisation strategy does result in better performance (see Section \ref{sec:results}).
For the reliability estimator, for each instance $\mathbf x_i$ we compare each annotation $a_{ij}$ to the labels predicted in the previous step.
If $a_{ij}$ is the same as the predicted label, we take the corresponding $r_{ij}$ to be $1$, and $0$ otherwise.
We then pre-train the reliability estimator $f_r$ to predict these values for $\mathbf r$.

\subsubsection*{EM Training}
\label{subsubsec:training}
We first consider training our model using expectation maximisation (EM;~\citeauthor{dempster1977maximum}~\citeyear{dempster1977maximum}).
This involves maximising the expectation of the complete log likelihood of the model with respect to the posterior of the latent variables in the model.
For the posterior of the model, we fix the parameters of the model and denote them $\boldsymbol \theta^{(k)}$ at iteration $k$ of the algorithm.
We only maximise the expectation with respect to the parameters $\boldsymbol \theta$ of the complete log likelihood.

The expectation is calculated as:
\begin{equation}
\small
\begin{split}
Q(\boldsymbol\theta \mid \boldsymbol\theta^{(k)}) &= \E[\log p(\mathbf{a,t,r \mid x}, \mat \theta)] \\
 &= \sum_{i=1}^N \E\left[\log p(t_i \mid \mathbf{x}_i, \mat \theta)\right] \\
 &+ \sum_{i=1}^N \sum_{j=1}^M \E\left[\log p(r_{ij} \mid \mathbf{x}_i, \mat \theta)\right] \\
& + \sum_{i=1}^N \sum_{j=1}^M \E\left[\log p(a_{ij} \mid t_i, r_{ij}, \mathbf{x}_i, \mat \theta)\right],
\label{equ:likelihood}
\end{split}
\end{equation}
where each expectation is calculated with respect to the posterior $p(\mat t, \mat r \mid \mat a, \mat x, \mat \theta^{(k)})$. 

\paragraph{E Step:} For the E step we compute the posterior with fixed parameters $\mat \theta^{(k)}$, $\pi_{ij}(t, r) = p(t_i = t, r_{ij} = r \mid \mat a_i, \mat x_i)$, as:
\begin{align}
\small
\begin{split}
&\pi_{ij}(t,r) = p(t_i = t, r_{ij} = r \mid \mat a_i, \mat x_i) \\
&\quad\propto p(t_i = t \mid \mat x_i) p(r_{ij} = r \mid \mat x_i) \\
&\quad\phantom{\propto} \cdot p(a_{ij} \mid t_i = t, r_{ij} = r, \mat x_i) \cdot \prod_{j' \ne j} \gamma_{ij'}^{(k)}(t)
\end{split} \label{equ:detailed_probability} \\
\small
\begin{split}
&\gamma_{ij}^k(t) = \sum_{r\prime \in \{0,1\}} \big(p(r_{ij} = r\prime \mid \mat x_i)\\
&\hspace{4em} \cdot p(a_{ij} \mid t_i = t, r_{ij} = r\prime, \mat x_i)\big),
\end{split} \label{equ:detailed_probability_2}
\end{align}
where we drop the dependency on $\mat \theta^{(k)}$ for brevity.
Note that $\pi_{ij}(t, r) = 0$ when $r = 1$ and $a_{ij} \ne t$.

We can then compute the marginalised posteriors, needed for \cref{equ:likelihood}, as follows:
\begin{align}
\small
p(t_i = t \mid \mat a_i, \mat x_i) &= \sum_{r \in \{0,1\}} \pi_{i1}(t, r) \label{equ:t_posterior}
\end{align}
\begin{align}
\small
p(r_{ij} = r \mid \mat a_i, \mat x_i) &=\sum_{t \in T} \pi_{ij}(t, r), \label{equ:r_posterior}
\end{align}
where the posterior $p(t_i, r_{i1} \mid \mat a_i, \mat x_i)$ of the model is chosen arbitrarily to marginalise over to get the posterior for $t_i$.

\paragraph{M Step:}
Using the posterior calculated in the E step we can compute the expectation of the complete log likelihood, $Q(\mat \theta \mid \mat \theta^{(k)})$, and calculate its gradient with respect to the parameters $\mat \theta$.
We then use gradient ascent to update the classifier and reliability estimator jointly.

\subsubsection*{Cross Entropy Training}
As an alternative training procedure, we also consider training the model using cross entropy.
As with expectation maximisation, we first calculate the posterior $\pi_{ij}(t, r)$ using the fixed parameters $\mat \theta^{(k)}$.
The networks $f_t$ and $f_r$ are then trained to minimise the cross entropy between the priors $p(t_i \mid \mat x_i)$ and $p(r_{ij} \mid \mat x_i)$, and the corresponding posteriors $p(t_i \mid \mat a_i, \mat x_i)$ and $p(r_{ij} \mid \mat a_i, \mat x_i)$.

The networks can be trained in an alternating fashion, with $f_r$ being trained while $f_t$ is kept fixed, and the other way around.
Denoting the parameters of $f_t$ as $\mat \theta_t$ and $f_r$ as $\mat \theta_r$, the loss functions for the respective networks then become
\begin{equation}
\small
\begin{split}
&L(\mat \theta_t \mid \mat \theta^{(k)}) = - \frac{1}{N} \sum_{i,t,r} \pi_{i1}(t, r) \log p(t_i \mid \mat x_i) \\
&L(\mat \theta_r \mid \mat \theta^{(k)}) = - \frac{1}{NM} \sum_{i,j,t,r} \pi_{ij}(t, r) \log p(r_{ij} \mid \mat x_i)
\end{split}
\normalsize
\end{equation}
Alternatively, they can be trained jointly by minimising the total cross entropy.
\begin{equation}
\small
L(\mat \theta \mid \mat \theta^{(k)}) = L(\mat \theta_t \mid \mat \theta^{(k)}) + L(\mat \theta_r \mid \mat \theta^{(k)})
\normalsize
\label{equ:total_cross_entropy}
\end{equation}
The training algorithm is summarised in \cref{alg:our_model}.
The algorithm is run until either a maximum number of iterations is reached, or the objective function stops improving.
\begin{algorithm}[t]
\begin{algorithmic}[1]
\scriptsize
\Require
  \Statex $\mat a$, the annotations
  \Statex $\mat x$, the instances
  \Statex $L$, the number of inner iterations
  \Statex $m$, the training mode: 0 for expectation maximisation, 1 for cross entropy (training alternatingly), 2 for cross entropy (training jointly)
  \Statex
  
  \State Pre-train $\mat \theta = \{\mat \theta_t, \mat \theta_r\}$ (Section \ref{subsubsec:initialisation})
  \State $k \gets 0$
  \While{stopping criteria not met}
    \State $k \gets k + 1$
    \State Calculate $\pi_{ij}(t, r)$ (Equation (\ref{equ:detailed_probability})) \Comment{E step}
    \If{$m = 0$} \Comment{M step}
      \ForTo{l}{1}{L}
        \State $\mat \theta \gets \mat \theta + \alpha \nabla_{\mat \theta} Q(\mat \theta \mid \mat \theta^{(k)})$
      \EndForTo
    \ElsIf{$m = 1$}
      \ForTo{l}{1}{L}
        \State $\mat \theta_r \gets \mat \theta_r - \alpha \nabla_{\mat \theta_r} L(\mat \theta_r \mid \mat \theta^{(k)})$
      \EndForTo
      \ForTo{l}{1}{L}
        \State $\mat \theta_t \gets \mat \theta_t - \alpha \nabla_{\mat \theta_t} L(\mat \theta_t \mid \mat \theta^{(k)})$
      \EndForTo
    \Else
      \ForTo{l}{1}{L}
      	\State $\mat \theta \gets \mat \theta - \alpha \nabla_{\mat \theta} L(\mat \theta \mid \mat \theta^{(k)})$
      \EndForTo
    \EndIf
  \EndWhile
\end{algorithmic}
\caption{Training procedure}
\label{alg:our_model}
\end{algorithm}
\section{Evaluation Settings}
\subsection{Data}
\label{subsec:data}
\subsubsection*{Simulated Annotators} 
\begin{figure}[t]
\centering
\includegraphics[scale=0.118]{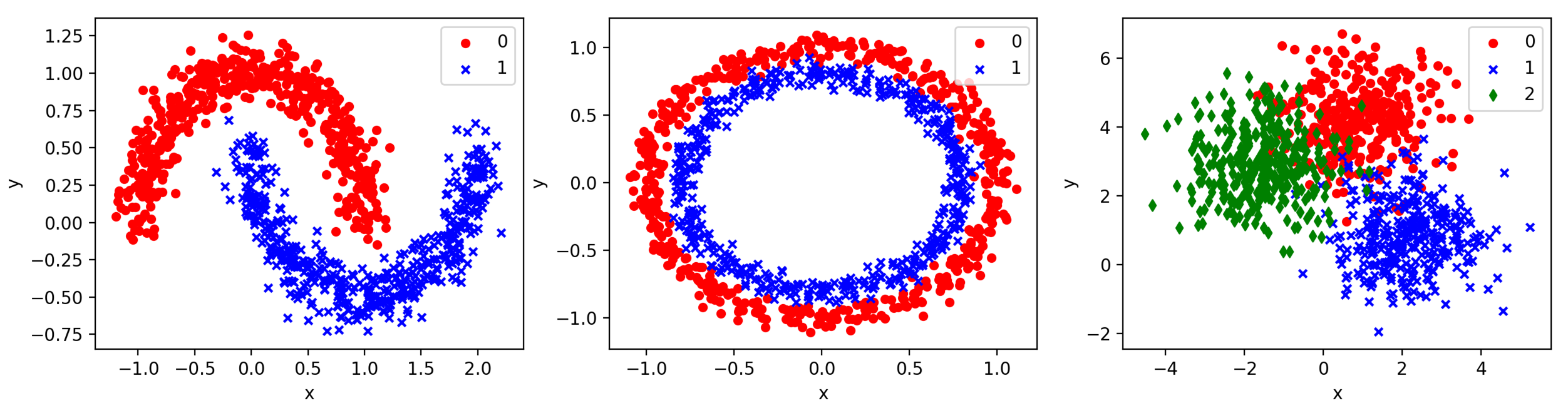}
\caption{Three 2-dimensional datasets.}
\label{fig:2d}
\end{figure}

\paragraph{2-Dimensional Datasets:}
In order to see whether our method can work well on simple cases, we create three 2-dimensional synthetic datasets, which we refer to as moon, circle and 3-class as shown in Figure \ref{fig:2d}.

\paragraph{Text Classification:}
For text classification we use the datasets Question Classification~\cite{li2002learning}, which contains short questions along with the type of answer expected, and Sentence Classification~\cite{chambers2013statistical}, which consists of sentences selected from medical publications. Examples of instance/class pairs for the text classification datasets include "Where is the Orinoco?" (class: "location") for the Question Classification dataset, and "New types of potent force clamps are discovered." (class: "author's own work") for the Sentence Classification dataset.

For these datasets that do not include crowd annotations, we synthesise annotations by simulating different annotators as follows: 1) \textbf{Narrow Expert}: has expertise in a single domain (i.e. class). For the instances of this class, the annotator will always provide the correct label. For other classes, a correct label will be provided with a probability of 0.65; otherwise, a random label will be selected with uniform probability; 2) \textbf{Broad Expert}: has expertise in every domain and only makes mistakes with a probability of 0.05; 3) \textbf{Random Annotator}: selects labels at random; 4) \textbf{Adversarial Annotator}: deliberately provides incorrect labels with a probability of 0.8.

For each of the datasets, we generated annotations using one narrow expert per class, one broad expert, one random annotator and one adversarial annotator, for a total of $|T| + 3$ annotators, where $|T|$ is the number of classes in the dataset.

In order to evaluate the generality of our model, we also apply it to another task in which we have 5 annotators with different overall reliabilities for the text classification tasks.
They produce incorrect labels with probabilities 0.1, 0.3, 0.5, 0.7, 0.9 respectively.
\subsubsection*{Real-World Crowdsourcing Annotators}
\paragraph{Recognising Textual Entailment:}
Finally, we evaluate our model on a real-world dataset for the recognising textual entailment (RTE) task~\cite{snow2008cheap}. Given a text pair, the annotator decides whether the hypothesis sentence can be inferred from the text fragment. The dataset includes both ground truth and crowdsourced labels from 164 annotators.

\Cref{tab:statistic_information} shows the number of instances of each class\footnote{The classes in the Sentence Classification dataset are defined as follows: AIMX---the goal of the paper; OWNX---the author's own work; CONT---the comparison including contrast and critique of past work; BASE---the past research that provides the basis for the work; MISC---any other sentences.} in the above-mentioned datasets.
\begin{table}[t]
\scriptsize
\centering
\begin{tabular}{clc}
\hline
\textbf{Dataset} & \textbf{Class} & \textbf{\# Instances} \\ \hline
moon& 0/1 & 500/500 \\
\hline
circle & 0/1 & 500/500 \\\hline
3-class & 0/1/2 & 334/333/333 \\
\hline
 & DESCRIPTION (DESC) & 1162 \\
 & ENTITY (ENTY) & 1250 \\
Question & ABBREV. (ABBR) & 86 \\
Classification & HUMAN (HUM) & 1223 \\
 & NUMERIC (NUM) & 896 \\
 & LOCATION (LOC)& 835 \\ \hline
 & AIMX & 94 \\
Sentence &  OWNX & 427 \\
Classification &  CONT & 104 \\
 & BASE & 33 \\ 
 & MISC & 852 \\ \hline
RTE & 0/1 & 400/400 \\ \hline
\end{tabular}
\caption{Classes and per-class instance counts.}
\label{tab:statistic_information}
\end{table}
\subsection{Experimental and Model Settings}
\label{subsec:settings}
\begin{table}[t]
\centering
\scriptsize
\begin{tabular}{cll}
\hline
\textbf{Datasets} & \textbf{Classifier} & \textbf{\# Units} \\ \hline
\begin{tabular}[c]{@{}c@{}}2-Dimensional\\ Datasets\\ (3 datasets) \end{tabular} & 2D$\rightarrow$FNN & 5 \\ \hline
\multirow{3}{*}{\begin{tabular}[c]{@{}c@{}}Text \\ Classification\\ (2 datasets)\end{tabular}} & BoW$\rightarrow$FNN & 100 \\
 & Avg. $\rightarrow$FNN & 50 \\
 & Embed.$\rightarrow$LSTM$\rightarrow$FNN & 100 \\ \hline
\multirow{2}{*}{RTE} & Cat. Avg. $\rightarrow$FNN & 100 \\
 & \begin{tabular}[c]{@{}l@{}}Embed.$\rightarrow$LSTM$\rightarrow$FNN\\ ~\cite{bowman2015large}\end{tabular} & 200 \\ \hline
\end{tabular}
\caption{Classifiers used in the experiments.}
\label{tab:classifiers}
\end{table}
Our model was implemented using the Chainer deep learning framework\footnote{\url{https://chainer.org/}}~\cite{tokui2015chainer}.

\paragraph{Classifier:}
As shown in \cref{tab:classifiers}, in each experiment the output of the classifier is generated by a feed-forward neural network (FNN).
Each FNN consists of an input layer, two hidden layers and a softmax output layer.
The number of hidden units in each layer is listed in the third column of the table.
The ReLU activation function~\cite{nair2010rectified} was applied after each hidden layer.
The output size of all the Long Short-Term Memory (LSTM;~\citeauthor{hochreiter1997long}, \citeyear{hochreiter1997long}) layers in our experiments is 100.

For the 2-dimensional classification task, each instance is simply represented using its position in 2-dimensional space.
For the text classification tasks, we investigated 3 methods of representing the sentences: bag-of-words (BoW) weighted by Term Frequency--Inverse Document Frequency (TFIDF), an average word embedding (Avg.) and the output at the last step of an LSTM layer (Embed.$\rightarrow$LSTM).
For the embedding we use word2vec embeddings pre-trained on Google News~\cite{mikolov2013distributed} for the question classification and RTE tasks, and a pre-trained embedding~\cite{8b865ebe417a41fbacb7241efe6a5490} trained on a combination of English Wikipedia, PubMed and PMC texts for the sentence classification task.

For the RTE task, we implemented two classifiers.
For the first one, each instance (i.e.\ a sentence pair) was represented as a concatenation of the average word embedding for each sentence (Cat. Avg.).
We also implemented ~\citeauthor{bowman2015large}~\shortcite{bowman2015large}, which runs each sentence through an LSTM, concatenates the outputs, and then feeds the concatenated output to an FNN with tanh activations.

\paragraph{Reliability Estimator:}
We model the reliability estimator as an FNN.
Its structure is the same as the classifier, albeit with different sizes of the two hidden layers.
For the experiments listed in \cref{tab:classifiers}, the number of units of each hidden layer in the FNN are 5, 100, 25, 25, 50, and 100 respectively.
The input to the estimator is the concatenation of the instance $\mathbf{x}_i$ (i.e. its original feature vector or the output of the last hidden layer of the classifier) and a one-hot vector representing the annotator identity.

\paragraph{Learning Settings:}
For every experiment we use the Adam~\cite{kingma2014adam} optimiser with a weight decay rate $0.001$, a gradient clipping of $5.0$, $\alpha=0.001$, $\beta_1=0.9$ and $\beta_2=0.999$.
We pre-train the classifier and reliability estimator for 200 epochs, using both majority voting and the model proposed by~\citeauthor{dawid1979maximum}~\shortcite{dawid1979maximum}.
The maximum number of outer iterations is set to 500 and 20 for EM training and cross entropy training respectively.
The number of inner iterations is 50 in both cases.

\paragraph{True Label Prediction and Reliability Estimation:}
After training, for each instance $\mat x_i$ we take its underlying label to be the most probable label according to the posterior of $t_i$ (see \cref{equ:t_posterior}).
We compared our predicted labels to the following state-of-the-art baselines: Majority Voting (MV), DS~\cite{dawid1979maximum},
GLAD~\cite{whitehill2009whose},
LFC~\cite{raykar2010learning},
CUBAM~\cite{welinder2010multidimensional},
~\citeauthor{yan2010modeling}~\shortcite{yan2010modeling},
KOS~\cite{NIPS2011_4396},
VI~\cite{NIPS2012_4627},
BCC~\cite{pmlr-v22-kim12}, 
MINIMAX~\cite{NIPS2012_4490},
MACE~\cite{hovy2013learning},
CATD~\cite{Li:2014:CAT:2735496.2735505}, PM~\cite{Li:2014:RCH:2588555.2610509} and EM-MV and Opt~\cite{zhang2014spectral}. 

Note that CUBAM,~\citeauthor{yan2010modeling}~\shortcite{yan2010modeling}, KOS and VI are only suitable for aggregating binary labels, and ~\citeauthor{yan2010modeling}~\shortcite{yan2010modeling} is the state-of-the-art method that models per-instance reliability. We take the reliability of annotator $j$ on instance $\mat x_i$ to be the posterior probability that $r_{ij}$ is 1 (see \cref{equ:r_posterior}).

\section{Results and Analysis}
\label{sec:results}
We measure the inter-annotator agreement (IAA) of each dataset. Fleiss's kappa~\cite{fleiss2013statistical}, denoted by $\kappa$, is measured for the 2-dimensional and text classification datasets, and Krippendorff's alpha~\cite{krippendorff1970estimating} is calculated for the RTE dataset\footnote{Although there are 164 annotators in total in this dataset, each instance was labelled by only 10 of these annotators. Therefore we use Krippendorff's alpha which is applicable to incomplete data to measure the inter-annotator agreement.}.
We find that the IAA values indicate slight agreement among annotators for all datasets.

Our experiments using different settings are shown as follows: our model is denoted by \textit{O}, with \textit{M} and \textit{D} denoting the model pre-trained using MV and DS respectively.
\textit{E} denotes training using expectation maximisation, while \textit{C} denotes cross entropy training. \textit{AL} and \textit{JT} denote cross entropy training done alternatingly and jointly, respectively. 

In the rest of this section, \cref{tab:f1_score_synthetic_toy,tab:f1_score_text,tab:stat_question_per_instance,tab:question_average_reliability,tab:stat_question_per_instance,tab:f1_score_different_classifiers} and \cref{tab:f1_score_real_world,tab:re_predict_performance,tab:reliabilities_on_real_world_dataset} present the results on the synthetic datasets and RTE dataset respectively. For the synthetic datasets, in \cref{tab:f1_score_synthetic_toy,tab:f1_score_text,tab:stat_question_per_instance,tab:question_average_reliability,tab:stat_question_per_instance}, we first consider a scenario where we have multiple narrow experts (N), one broad expert (B), one random annotator (R) and one adversarial annotator (A). In \cref{tab:f1_score_different_classifiers}, we further consider a scenario with 5 annotators, each of differing reliability, as explained in Section \ref{subsec:data}.

\Cref{tab:f1_score_synthetic_toy} shows that our method performs well on the 2-dimensional datasets, obtaining higher label prediction F1 scores than the baselines. We omit the analysis of the true label prediction and reliability estimation results on these datasets as all models performed similarly,
choosing instead to focus the discussion on the results for the NLP tasks.
\renewcommand{\tabcolsep}{1.5mm}
\begin{table}
\centering
\resizebox{\columnwidth}{!}{%
\begin{tabular}{lccc}
\hline
& \multicolumn{3}{c}{\textbf{2-Dimensional Datasets}} \\\cline{2-4} 
& \textbf{moon} & \textbf{circle} & \textbf{3-class} \\
& $\kappa=0.029$&$\kappa=0.029$ &$\kappa=0.153$ \\\hline
MV & 84.6 & 84.6 & 89.0 \\
DS~\cite{dawid1979maximum} & 97.8 & 97.9 & 99.4 \\
GLAD~\cite{whitehill2009whose} & 97.9 & 97.9 & 93.7 \\
LFC~\cite{raykar2010learning} & 97.9 & 97.9 & 99.4 \\
CUBAM~\cite{welinder2010multidimensional} & 97.9 & 97.9 & - \\
Yan et al.~\cite{yan2010modeling} & 84.1 & 84.1 & - \\
KOS~\cite{NIPS2011_4396} & 96.7 & 96.7 & - \\
VI~\cite{NIPS2012_4627} & 97.9 &97.9 & - \\
BCC~\cite{pmlr-v22-kim12} & 97.9 & 97.9 & 99.4 \\
MINIMAX~\cite{NIPS2012_4490} & 97.9 & 97.9 & 99.2 \\
MACE~\cite{hovy2013learning} & 97.8 & 97.9 & 97.0  \\
CATD~\cite{Li:2014:CAT:2735496.2735505} & 94.5 & 94.5 & 94.3 \\
PM~\cite{Li:2014:RCH:2588555.2610509} & 94.5 & 94.5 & 94.3 \\
EM-MV~\cite{zhang2014spectral} & 97.9 & 97.9 & 96.3 \\
EM-Opt~\cite{zhang2014spectral} & 72.7 & 72.7 & 93.2 \\
\hline
O-ME & 97.0 & 97.7 & 94.8  \\
O-MC & 97.3 & 98.9 & 99.3  \\
O-DE & 91.2 & 97.2 & 96.5 \\
O-DC-AL & 99.7 & \textbf{99.7} & \textbf{99.5}\\
O-DC-JT & \textbf{99.8} & 99.6 & \textbf{99.5}\\
\hline
\end{tabular}
}
\caption{F1 scores of predicted labels on the 2-dimensional datasets when using the output of the last hidden layer of the classifier to represent an instance for the reliability estimator.}
\label{tab:f1_score_synthetic_toy}
\end{table}
\subsection{Classifier and Reliability Estimator}
In order to explore the separate performance contribution of classifier and reliability estimator, we compare the performance of our model to a classifier pre-trained using DS labels, as well as a variant of our model without the reliability estimator, i.e. setting all the annotators have the same reliability on all the instances.

As shown in \cref{tab:f1_score_text,tab:f1_score_different_classifiers,tab:f1_score_real_world}, the pre-trained classifier performed worse than some aggregation methods.
This indicates that the noise in the labels predicted by DS has an adverse effect on the training of the classifier.
The much lower performance of the model with the reliability estimator removes hints at the importance of modelling per-annotator reliability to ensure accurate predictions.

\subsection{Instance Representation for Reliability Estimator}
For the representation of the instance $\mathbf{x}_i$ as it is fed to the reliability estimator, we compared the performance of using the original feature vector of $\mathbf{x}_i$ to using the last hidden layer output of the classifier (which we refer to as the ``full model'').

We found that using the hidden layer representation can not only improve the label prediction performance (see \cref{tab:f1_score_text,tab:f1_score_different_classifiers,tab:f1_score_real_world}), but also sped up the training compared to using the feature vector directly.
The hidden layer representation allows us to reduce the number of parameters in the model, by sharing parameters with the classifier.

\subsection{Full Model on Synthetic Datasets}
Based on the results of the full model in \cref{tab:f1_score_text}, we can conclude that per-instance reliability modelling is beneficial to the label prediction task, and using the average pre-trained embedding can result in slightly better performance. 
It is worth noting that the method used to pre-train the model had a noticeable effect on its performance, with better F1 scores being obtained when using DS pre-training.
In the following experiments we only consider models pre-trained using the DS algorithm.

\renewcommand{\tabcolsep}{1.5mm}
\begin{table}[t!]
\centering
\scriptsize
\begin{tabular}{lcccc}
\hline
\multirow{2}{*}{\textbf{}} & \multicolumn{2}{c}{\textbf{\begin{tabular}[c]{@{}c@{}}Question\\ Classification\end{tabular}}} & \multicolumn{2}{c}{\textbf{\begin{tabular}[c]{@{}c@{}}Sentence\\ Classification\end{tabular}}} \\ 
& \multicolumn{2}{c}{$\kappa=0.094$} & \multicolumn{2}{c}{$\kappa=0.0634$} \\
\cline{2-5}
& \textbf{BoW} & \textbf{Avg.} & \textbf{BoW} & \textbf{Avg.} \\\hline
MV &  78.8 & 78.8 & 71.3 & 71.3 \\
DS~\cite{dawid1979maximum}  & 98.3 & 98.3 & 97.3 & 97.3 \\
GLAD~\cite{whitehill2009whose} & 87.1 &87.1 & 79.9 & 79.9 \\
LFC~\cite{raykar2010learning} & 98.2 &98.2 & 97.0 &97.0 \\
BCC~\cite{pmlr-v22-kim12} & 98.3 & 98.3 & 98.1 & 98.1 \\
MINIMAX~\cite{NIPS2012_4490} & 28.2 &28.2 & 50.9 & 50.9 \\
MACE~\cite{hovy2013learning}  & 91.6 & 91.6 & 63.6 & 63.6 \\
CATD~\cite{Li:2014:CAT:2735496.2735505} & 91.1 & 91.1 & 92.2 &92.2 \\
PM~\cite{Li:2014:RCH:2588555.2610509} & 91.1 & 91.1 & 92.2 & 92.2 \\
EM-MV~\cite{zhang2014spectral} & 88.3 &88.3 & 66.9 &66.9 \\
EM-Opt~\cite{zhang2014spectral} & 13.5 & 13.5 & 26.5 & 26.5 \\
\hline
Classifier (pre-trained by DS labels) & 92.3 & 86.5& 89.6&85.4\\\hline
O-DE (without reliability estimator) & 92.1 & 91.2 & 77.6 & 87.1 \\
O-DC-AL (without reliability estimator) & 90.5 & 93.2 & 84.0 & 88.2 \\
O-DC-JT (without reliability estimator) & 88.7 & 91.1 & 78.5 & 87.4 \\\hline
O-DE (using feature vector) & 95.1 & 97.0 & 88.9 & 97.8 \\
O-DC-AL (using feature vector) & 97.5 & 98.9 & 97.2 & 97.5 \\
O-DC-JT (using feature vector) & 97.5 & 98.9 & 97.2 & 97.5 \\\hline
O-ME (full model) & 92.4 & 95.0 & 84.4 & 89.2 \\
O-MC (full model) & 97.2 & 98.3 & 90.6 & 94.6 \\
O-DE (full model) & 98.0 & 98.3 & \textbf{97.7} & \textbf{98.9} \\
O-DC-AL (full model) & 98.3 & 98.6 & 97.6 & 97.9 \\
O-DC-JT (full model) & \textbf{98.7} & \textbf{99.0} & 97.3 & 97.8 \\
\hline
\end{tabular}
\caption{F1 scores of predicted labels on the text classification datasets.}
\label{tab:f1_score_text}
\end{table}

\begin{table}
\centering
\resizebox{0.75\columnwidth}{!}{%
\begin{tabular}{cccccccc}
\hline
\multirow{2}{*}{} & \multicolumn{7}{c}{\textbf{Question Classification}} \\ \cline{2-8}
 & \textbf{DESC} & \textbf{ENTY} & \textbf{ABBR} & \textbf{HUM} & \textbf{NUM} & \textbf{LOC} & \textbf{Accuracy} \\\hline
1 (N) & \cellcolor{gray!50}99 & 1 & 0 & 0 & 0 & 0 & 100 \\
2 (N)& 0 & \cellcolor{gray!50}100 & 0 & 0 & 0 & 0 & 100 \\
3 (N)& 31 & 13 & \cellcolor{gray!50}40 & 5 & 2 & 9 & 100 \\
4 (N)& 0 & 0 & 0 & \cellcolor{gray!50}100 & 0 & 0 & 100 \\
5 (N)& 0 & 0 & 0 & 0 & \cellcolor{gray!50}100 & 0 & 100 \\
6 (N)& 0 & 0 & 0 & 0 & 0 & \cellcolor{gray!50}100 & 100 \\
7 (B)& \cellcolor{gray!50}20 & \cellcolor{gray!50}0 & \cellcolor{gray!50}0 & \cellcolor{gray!50}77 & \cellcolor{gray!50}0 & \cellcolor{gray!50}3 & 100 \\
8 (R)& 30 & 32 & 8 & 8 & 15 & 7 & 100 \\
9 (A)& 45 & 19 & 9 & 14 & 5 & 8 & 100 \\ \hline
\end{tabular}
}
\caption{Number of correctly labelled examples for each annotator (N: narrow expert, B: broad expert, R: random annotator and A: adversarial annotator) among the 100 instances with highest per-instance reliability on the question classification dataset.}
\label{tab:stat_question_per_instance}
\end{table}

\begin{table}[t!]
\centering
\resizebox{0.76\columnwidth}{!}{%
\begin{tabular}{cccccccccccc}
\hline
\multirow{2}{*}{} & \multicolumn{7}{c}{\textbf{Question Classification}} \\ \cline{2-8}
 & \textbf{DESC} & \textbf{ENTY} & \textbf{ABBR} & \textbf{HUM} & \textbf{NUM} & \textbf{LOC}  & \textbf{Overall}\\\hline
1 (N) & \cellcolor{gray!50}90.6 & 24.4 & 35.5 & 21.4 & 26.6 & 22.2 & 36.8 \\
2 (N)& 28.7 & \cellcolor{gray!50}92.3 & 26.6 & 25.2 & 24.3 & 26.7&37.3 \\
3 (N)& 24.7 & 24.9 & \cellcolor{gray!50}75.8 & 23.5 & 24.7 & 23.6 & 32.9 \\
4 (N)& 27.5 & 26.3 & 32.3 & \cellcolor{gray!50}93.4 & 24.2 & 23.7 & 37.9 \\
5 (N)& 25.6 & 21.9 & 25.2 & 25.7 & \cellcolor{gray!50}93.8 & 26.0 & 36.4\\
6 (N)& 26.4 & 25.0 & 24.1 & 22.1 & 27.5 & \cellcolor{gray!50}94.4 & 36.6\\
7 (B)& \cellcolor{gray!50}94.3 & \cellcolor{gray!50}93.5 & \cellcolor{gray!50}95.1 & \cellcolor{gray!50}94.2 & \cellcolor{gray!50}93.4 & \cellcolor{gray!50}93.9 & 94.2\\
8 (R)& 5.60 & 6.30 & 8.40 & 4.80 & 6.20 & 5.90 & 6.20\\
9 (A)& 9.10 & 8.90 & 12.0 & 8.20 & 7.90 & 8.30 & 9.10 \\\hline
\end{tabular}
}
\caption{Average reliability of each annotator among the 100 instances with the highest per-instance reliability on the question classification dataset.}
\label{tab:question_average_reliability}
\end{table}

In order to investigate whether our method can successfully capture per-instance annotator reliability, for each annotator,
we counted the number of correctly labelled instances and calculated the average reliability for each class among the top 100 instances with the highest per-instance reliability as shown in \cref{tab:stat_question_per_instance} and \ref{tab:question_average_reliability}\footnote{We omit the results for the sentence classification task for lack of space, as we consider the results on the question classification dataset to be representative.}.
The cells with grey background colour indicate which domain, or class, the annotator has expertise in. It can be seen that all annotators obtain high accuracy on these instances. In general our method also captured the varying expertise of each narrow annotator, estimating their reliability on instances belonging to the corresponding classes as particularly high.

For these experiments in \Cref{tab:f1_score_different_classifiers}, we also investigated the performance when using two different classification models. As seen in this table, both of them outperformed all baselines significantly.

\begin{table}[t]
\centering
\scriptsize
\resizebox{1.0\columnwidth}{!}{%
\begin{tabular}{lcccc}
\hline
\multirow{2}{*}{} & \multicolumn{2}{c}{\textbf{\begin{tabular}[c]{@{}c@{}}Question\\ Classification\end{tabular}}} & \multicolumn{2}{c}{\textbf{\begin{tabular}[c]{@{}c@{}}Sentence\\ Classification\end{tabular}}} \\
& \multicolumn{2}{c}{$\kappa=0.126$} & \multicolumn{2}{c}{$\kappa=0.0776$} \\
\cline{2-5}
 & \textbf{FNN} & \textbf{LSTM+FNN} & \textbf{FNN} & \textbf{LSTM+FNN} \\\hline
MV & 71.8 & 71.8 & 65.8 & 65.8 \\
DS~\cite{dawid1979maximum} & 90.1 & 90.1 & 83.9 & 83.9 \\
GLAD~\cite{whitehill2009whose} & 80.9 & 80.9 & 71.8 & 71.8 \\
LFC~\cite{raykar2010learning} & 88.3 & 88.3 & 80.3 & 80.3 \\
BCC~\cite{pmlr-v22-kim12} & 90.4 & 90.4 & 85.6 & 85.6 \\
MINIMAX~\cite{NIPS2012_4490} & 30.4 & 30.4 & 44.0 & 44.0 \\
MACE~\cite{hovy2013learning} & 84.6 & 84.6 & 62.4 & 62.4 \\
CATD~\cite{Li:2014:CAT:2735496.2735505} & 85.2 & 85.2 & 80.5 & 80.5 \\
PM~\cite{Li:2014:RCH:2588555.2610509} & 85.2 & 85.2 & 80.5 & 80.5 \\
EM-MV~\cite{zhang2014spectral} & 75.4 & 75.4 & 56.6 & 56.6 \\
EM-Opt~\cite{zhang2014spectral} & 19.4 & 19.4 & 14.4 & 14.4 \\
\hline
Classifier (pre-trained by DS labels) & 89.7 & 76.6& 79.6&78.7\\\hline
O-DE (without reliability estimator) & 91.9 & 91.0 & 80.1 & 82.5 \\
O-DC-AL (without reliability estimator) & 91.8 & 91.2 & 80.0 & 83.5 \\
O-DC-JT (without reliability estimator) & 91.5 & 91.6 & 79.3 & 83.5\\\hline
O-DE (using feature vector) & 94.3 & - & 81.7 & - \\
O-DC-AL (using feature vector) & 94.1 & - & 86.2 & - \\
O-DC-JT (using feature vector)& 94.1 & - & 86.2 & -\\\hline
O-DE (full model)& 94.5 & 94.1 & 86.1 & 86.1 \\
O-DC-AL (full model)& 94.7 & \textbf{96.0} & \textbf{90.3} & 88.5\\
O-DC-JT (full model)& \textbf{95.1} & 95.1 & 88.5 & \textbf{89.0}\\
\hline
\end{tabular}
}
\caption{F1 scores on text classification tasks when only the reliability differs between the annotators.}
\label{tab:f1_score_different_classifiers}
\end{table}


\renewcommand{\tabcolsep}{1.5mm}
\subsection{Full Model on RTE Dataset}
\Cref{tab:f1_score_real_world} presents the label prediction performance on the RTE dataset.
As not every annotator has provided labels for every instance in this dataset, for both the EM and cross entropy training we simply omitted missing instance/annotator pairs when calculating the loss functions.
As seen in the table, most of the baselines obtained high performance as the textual entailment recognition task is easy for non-expert annotators. However, our full model still achieved better prediction performance than all of the baseline methods.
\begin{table}[t]
\centering
\scriptsize
\resizebox{0.75\columnwidth}{!}{%
\begin{tabular}{lccc}
\hline
\textbf{Method} & \textbf{F-measure} \\\hline
MV &  91.9 \\
DS~\cite{dawid1979maximum} &  92.6 \\
GLAD~\cite{whitehill2009whose} & 92.4 \\
LFC~\cite{raykar2010learning} & 92.5 \\
CUBAM~\cite{welinder2010multidimensional} &  92.6 \\
Yan et al.~\cite{yan2010modeling} &  90.4 \\
KOS~\cite{NIPS2011_4396} & 63.2 \\
VI~\cite{NIPS2012_4627} & 92.5  \\
BCC~\cite{pmlr-v22-kim12} &  92.3 \\
MINIMAX~\cite{NIPS2012_4490} &92.4  \\
MACE~\cite{hovy2013learning} &  92.4 \\
CATD~\cite{Li:2014:CAT:2735496.2735505} &  92.3 \\
PM~\cite{Li:2014:RCH:2588555.2610509} &  92.0 \\
EM-MV~\cite{zhang2014spectral} & 92.5  \\
EM-Opt~\cite{zhang2014spectral} & 92.4  \\
\hline
Classifier (pre-trained by DS labels) & 89.9 \\\hline
O-DC-JT (FNN) (without reliability estimator)& 90.2 \\
\hline
O-DC-JT (FNN) (using feature vector) & 92.6 \\\hline
O-DE (FNN) (full model) &  92.9 \\
O-DC-AL (FNN) (full model) & 92.8 \\
O-DC-JT (FNN) (full model)&  \textbf{93.0} \\
O-DE (LSTM+FNN) (full model)&  92.7 \\
O-DC-AL (LSTM+FNN) (full model)& 92.8 \\
O-DC-JT (LSTM+FNN) (full model)& 92.8 \\
\hline
\end{tabular}
}
\caption{Performance of predicted labels on the RTE dataset (Krippendorff's alpha $=0.0995$).}
\label{tab:f1_score_real_world}
\end{table}


We also investigated the effectiveness of our model for removing noisy labels.
We compare our model to the five best-performing baselines (DS, LFC, CUBAM, VI and EM-MV in \cref{tab:f1_score_real_world}).
Each of these models are trained on the RTE dataset, after which the least reliable annotation for each instance is removed.
We use the per-instance reliability for our model, the global reliability score of each annotator for LFC, CUBAM and VI, and the per-category annotator reliability for DS and EM-MV as the measure of the reliability of each annotation.
For each of these models, we then retrain the models in \cref{tab:f1_score_real_world} using the denoised dataset; the difference in performance can be seen in \cref{tab:re_predict_performance}. We can see that using per-instance reliability results in the largest improvement, while only considering the annotators' overall reliability may cause a reduction in performance.

In order to analyse the per-instance reliability of the human annotators, for each annotator we rank the instances according to the annotator's per-instance reliability.
We look at the top 15 and bottom 15 instances, then count how many of them were correctly labelled (Cor. Labels) as well as the average reliability on these instances (Avg. Reliability). \Cref{tab:reliabilities_on_real_world_dataset} shows the results of five annotators\footnote{For lack of space, we only present the results for 5 of the 164 annotators.}. It can be seen that each annotator has considerably different reliabilities across instances.
\begin{table}[t!]
\centering
\resizebox{\columnwidth}{!}{%
\begin{tabular}{lcccccc}
\hline
\textbf{Method} & \textbf{LFC} & \textbf{CUBAM} & \textbf{VI} & \textbf{DS} & \textbf{EM-MV} & \textbf{Ours}\\ \hline
MV &  -0.2 & -0.9 & +0.6 & -0.2 &-0.2 & \textbf{+0.8} \\
DS~\cite{dawid1979maximum} &  0 & -0.2 &+0.1&0 &0&\textbf{+0.6}\\ 
GLAD~\cite{whitehill2009whose} & \textbf{+0.2}  & 0 &0&\textbf{+0.2} &-0.3&\textbf{+0.2}\\
LFC~\cite{raykar2010learning} &+0.1 &-0.1&+0.3&+0.1 &+0.1&\textbf{+0.6} \\
CUBAM~\cite{welinder2010multidimensional} &  +0.3 & 0 &+0.4&+0.3 &-0.2&\textbf{+0.9}\\
Yan et al.~\cite{yan2010modeling} & +1.5& +0.4  &+2.3&+1.5 & +0.8&\textbf{+2.5}\\
KOS~\cite{NIPS2011_4396} & +0.7 &+3.9 &+11.3&+0.7&+13.4&\textbf{+17.3}\\
VI~\cite{NIPS2012_4627} &+0.1 &+0.1&+0.1&+0.1 &+0.2&\textbf{+0.6} \\
BCC~\cite{pmlr-v22-kim12} &+0.3 &+0.2&+0.3&+0.3 &+0.4&\textbf{+0.8} \\
MINIMAX~\cite{NIPS2012_4490} &+0.3 &-0.4&\textbf{+0.7}&+0.3 &+0.2&\textbf{+0.7} \\
MACE~\cite{hovy2013learning} &  +0.2 & -0.2&0&+0.2 &0&\textbf{+0.4} \\
CATD~\cite{Li:2014:CAT:2735496.2735505}&+0.3 &-0.8 &-0.2&+0.3 &+0.2&\textbf{+0.3}\\
PM~\cite{Li:2014:RCH:2588555.2610509} &\textbf{+0.9} &0&+0.3&\textbf{+0.9} &+0.1&\textbf{+0.9} \\
EM-MV~\cite{zhang2014spectral} &+0.7 &+0.6&+0.7&+0.7 &+0.7&\textbf{+1.1} \\
EM-Opt~\cite{zhang2014spectral} &+0.2 &+0.5&+0.2&+0.2 &+0.6&\textbf{+0.7} \\
O-DC-JT (FNN) (full model) &  +0.1 &0  &+0.3&+0.1 &+0.1&\textbf{+0.5}\\
\hline
\end{tabular}
}
\caption{F1 score improvements after removing the label produced by the least reliable annotator by using the estimated overall reliability (LFC, CUBAM, VI, DS, EM-MV) and per-instance reliability (Ours).}
\label{tab:re_predict_performance}
\end{table}
\begin{table}[t!]
\resizebox{\columnwidth}{!}{%
\begin{tabular}{ccccccc}
\hline
 &  & \multicolumn{2}{c}{\textbf{True Entailment}} & \multicolumn{2}{c}{\textbf{False Entailment}} & \textbf{} \\
 & \textbf{Annotator} & \textbf{\#Cor. Labels} & \textbf{Avg. Reliability} & \textbf{\#Cor. Labels} & \textbf{Avg. Reliability} & \textbf{Acc.} \\ \hline
\multirow{5}{*}{\begin{tabular}[c]{@{}c@{}}Top\\ 15 \\ Instances\end{tabular}} & 1 & 15 & 95.5 & - & - & 100 \\
 & 2 & 15 & 92.6 & - & - & 100 \\
 & 3 & 10 & 88.4 & 3 & 86.9 & 86.7 \\
 & 4 & 11 & 94.2 & 2 & 92.6 & 86.7 \\
 & 5 & 8 & 71.9 & 2 & 23.7 & 66.7 \\ \hline
\multirow{5}{*}{\begin{tabular}[c]{@{}c@{}}Bottom\\ 15 \\ Instances\end{tabular}} & 1 & 4 & 0.1 & 2 & 0.1 & 40 \\
 & 2 & 1 & 1.0 & 8 & 51.4 & 60 \\
 & 3 & 0 & 9.7 & 1 & 13.8 & 6.6 \\
 & 4 & 2 & 28.6 & 5 & 63.1 & 46.7 \\
 & 5 & 3 & 22.3 & 2 & 23.7 & 33.3 \\ \hline
\end{tabular}
}
\caption{Number of correct labels and average reliability for each annotator among the instances with highest and lowest per-instance reliability on the RTE dataset.}
\label{tab:reliabilities_on_real_world_dataset}
\end{table}

\subsection{Training Stability}
\paragraph{Pre-training:} 
As discussed in Section \ref{subsec:learning}, the predicted labels produced by a simpler method are used for pre-training. Although these labels are not perfect, we assume that our method can still learn some useful information from them for a better starting point than random parameter initialisation. 

\paragraph{EM and Cross Entropy Training:} From \cref{tab:f1_score_synthetic_toy,tab:f1_score_text,tab:f1_score_different_classifiers,tab:f1_score_real_world}, it can be seen that, in most cases, using cross entropy achieved much better and more stable performance than the models learned using EM training. We also noticed that the objective function would improve when using cross entropy training, and tended to converge faster in our experiments---generally within just a few epochs. Therefore, we recommend to use this training method in practice.

\paragraph{Early Stopping:} When using both EM and cross entropy training, we found that even if the objective function improved between iterations, the label prediction performance would eventually start to decrease. It is worth to investigate the reason for this phenomenon. To counteract this issue we used early stopping, where training is halted when the objective function does not improve more than 0.001 between iterations. Another option is to reduce the maximum number of outer iterations, e.g. to 20.

\section{Conclusion and Future Work}
We propose a novel probabilistic model which learns from noisy labels produced by multiple annotators for NLP crowdsourcing tasks by incorporating a classifier and a reliability estimator. Our work constitutes the first effort to model the per-instance reliability of annotators for both binary and multi-class NLP labelling tasks.
We investigate two methods of training our model using the EM algorithm and cross entropy. Experimental results on 6 datasets including synthetic and real datasets demonstrate that our method can not only capture the per-instance reliability of each annotator, but also obtain better label prediction and the least reliable annotator detection performance compared to state-of-the-art baselines.

For future work, we plan to apply our model to other NLP tasks such as relation extraction and named entity recognition.
We also plan to investigate the use of variational inference~\cite{Jordan1999} as a means of training our model. 
Using variational inference might improve the stability and performance of our model. 

\section*{Acknowledgement}
We would like to thank the anonymous reviewers and Paul Thompson for their valuable comments. Discussions with Austin J. Brockmeier have been insightful. The work is funded by \textit{School of Computer Science Kilburn Overseas Fees Bursary} from University of Manchester.

\bibliography{naaclhlt2019}
\bibliographystyle{acl_natbib}
\end{document}